\documentclass[fleqn,10pt,twocolumn]{SICE14_fixed}

\title{Decision Maker using Coupled Incompressible-Fluid Cylinders}

\author{Song-Ju Kim${}^{1}$ and Masashi Aono${}^{2}$}
\speaker{Song-Ju Kim}

\affils{${}^{1}$WPI Center for MANA, National Institute for Materials Science, Tsukuba, Japan\\
${}^{2}$Earth-Life Science Institute, Tokyo Institute of Technology, Tokyo \& PRESTO JST, Japan\\
(E-mail: KIM.Songju@nims.go.jp)\\
}
\abstract{%
The multi-armed bandit problem (MBP) is the problem of finding, as accurately and quickly as possible, the most profitable option from a set of options that gives stochastic rewards by referring to past experiences.
Inspired by fluctuated movements of a rigid body in a tug-of-war game, we formulated a unique search algorithm that we call the `tug-of-war (TOW) dynamics' for solving the MBP efficiently~\cite{kim1,kim2,kim3,kim4,kim5}.
The cognitive medium access, which refers to multi-user channel allocations in cognitive radio, can be interpreted as the competitive multi-armed bandit problem (CMBP); the problem is to determine the optimal strategy for allocating channels to users which yields maximum total rewards gained by all users~\cite{kim6}.
Here we show that it is possible to construct a physical device for solving the CMBP, which we call the `TOW Bombe', by exploiting the TOW dynamics existed in coupled incompressible-fluid cylinders.
This analog computing device achieves the `socially-maximum' resource allocation that maximizes the total rewards in cognitive medium access without paying a huge computational cost that grows exponentially as a function of the problem size.
}

\begin{document}

\maketitle


\section*{Introduction}

Consider two slot machines.
Both machines have individual reward probabilities $P_A$ and $P_B$.
At each trial, a player selects one of machines and obtains some reward, for example, a coin, with the corresponding probability.
The player wants to maximize the total reward sum obtained after a certain number of selections.
However, it is supposed that the player does not know these probabilities.
The multi-armed bandit problem (MBP) is to determine the optimal strategy for selecting the machine which yields maximum rewards by referring to past experiences.

In our previous studies~\cite{kim1,kim2,kim3,kim4,kim5,kim6}, we have shown that our proposed algorithm called the Tug-of-War (TOW) dynamics is more efficient than other well-known algorithms such as the modified $\epsilon$-greedy algorithm and modified softmax algorithm, and comparable to the `upper confidence bound1-tuned (UCB1T) algorithm' that is known as the best algorithm among parameter-free algorithms~\cite{auer}. 
Moreover, the TOW dynamics effectively adapts to a changing environment in which the reward probabilities dynamically switch.
The algorithms for solving the MBP are useful for various applications, such as the cognitive radio~\cite{cog,cog2}, web advertising~\cite{web}, and the Monte-Carlo tree search that is used for programming computers to play `game of GO'~\cite{uct,mogo}.

Recently, the cognitive medium access problem is one of the hottest topics in the field of mobile communications~\cite{cog,cog2}.
The underlying idea is to allow unlicensed users (i.e., cognitive users) to access the available spectrum when the licensed users (i.e., primary users) are not active. 
The cognitive medium access is a new medium access paradigm in which the cognitive users should not interfere with the licensed users.
To avoid interfering with the primary network, the cognitive users must first probe to determine whether there are primary activities in each channel before transmission.

\begin{figure}[t]
\centering
\includegraphics[height=30mm]{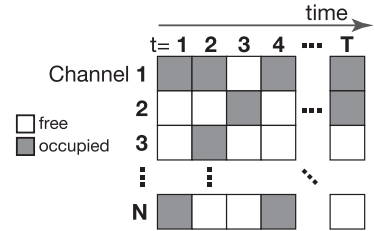}
\caption{Channel model.}
\label{fig:channel}
\end{figure}

Figure~\ref{fig:channel} shows the channel model proposed by Lai et al.~\cite{cog,cog2}. 
There is a primary network consisting of $N$ channels, each with bandwidth B. 
The users in the primary network are operated in a synchronous time-slotted fashion. 
It is assumed that, at each time slot, channel $i$ is free with probability $P_i$. 
The cognitive users do not know $P_i$ a priori.
At each time slot, the cognitive users attempt to exploit the availability of channels in the primary network by sensing the activity in this channel model.
In this setting, a single cognitive user can access only a single channel at any given time. 
The problem is to derive an optimal accessing strategy for choosing channels that maximizes the expected throughput obtained by the cognitive user. 
This situation can be interpreted as the multi-user competitive bandit problem (CMBP).

For simplicity, we consider the minimum CMBP, i.e., 2 cognitive (unlicensed) users (1 and 2) and 2 channels ($A$ and $B$).
Each channel is not occupied by primary (licensed) users with the probability $P_i$.
In the MBP context, we assume that the user accessing a free channel can get some reward, for example a coin, with the probability $P_i$.
Table~\ref{table:1} shows the payoff matrix for user 1 and 2.
\begin{table}[b]
\caption{Payoff matrix for user 1 (user 2).}
\label{table:1}
\begin{center}
\begin{tabular}{|c|c|c|} \hline \hline
   & user 2: A & user 2: B \\ \hline
user 1: A & $P_A/2$ ($P_A/2$) & $P_A$ ($P_B$) \\ \hline
user 1: B  & $P_B$ ($P_A$) & $P_B/2$ ($P_B/2$) \\\hline
\end{tabular}
\end{center}
\end{table}
When two cognitive users select the same channel, the collision occurs, and the reward is evenly split between the collided users.

In order to develop a unified framework for the design of efficient, and low complexity, cognitive medium access protocols, we have to seek an algorithm that can obtain the maximum total rewards (scores) in the CMBP context.
In order to acquire the maximum total rewards, the algorithm has to have a mechanism that can avoid the `Nash equilibrium' which is the natural consequence for a group of independent selfish users.

In this study, we demonstrate that overall optimization (the maximum total rewards) can be derived by using a physical device consisting of two kinds of incompressible-fluid in two or more cylinders.  
We call this analog computing device the `Tug-of-War (TOW) Bombe' because it is analogous to the `Turing Bombe,' which is an analog electric circuit developed by the British army during World War II for decoding the `enigma code' of the German army~\cite{tbom}. 
If one tries to solve the CMBP for $M$ users and $N$ channels using a conventional digital computer, it is necessary to calculate evaluation values of $O(N^M)$ for each iteration; the computational cost for solving the CMBP grows as an exponential function of $N$ and $M$.
Nevertheless, the TOW Bombe enables to solve the problem without paying the exponential computational cost. 
At each iteration, the TOW Bombe only requires $M$ up-and-down operations for controlling the fluid interface levels in the corresponding cylinders. 

\section{The Tug-of-War Dynamics}

In the previous studies~\cite{kim4,kim6}, we proposed the Tug-of-War (TOW) dynamics.
\begin{figure}[b]
\centering
\includegraphics[height=28mm]{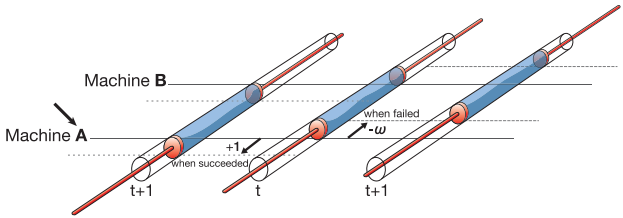}
\caption{TOW dynamics.}
\label{fig:stow}
\end{figure}
Consider incompressible-fluid in a cylinder, as shown in Fig.~\ref{fig:stow}.
Here, variable $X_k$ corresponds to the displacement of terminal $k$ from an initial position, where $k\in \{A,B\}$.
If $X_k$ is greater than $0$, we consider that the liquid selects machine $k$.

We used the following estimate $Q_k$ ($k\in \{A,B\}$):
\begin{equation}
Q_k(t)  =   N_k(t) - (1 + \omega) L_k(t). \label{qq}
\end{equation}
Here, $N_k$ is the number of playing machine $k$ until time $t$, and $L_k$ is the number of non-rewarded (i.e., failed) events in $k$ until time $t$, where $\omega$ is a weighting parameter.

The displacement $X_A$ ($= - X_B$) is determined by the following difference equation:
\begin{eqnarray}
X_A(t) & = & Q_A(t) - Q_B(t) + \delta .\label{eq:os}
\end{eqnarray}
Here, $\delta(t)$ is an arbitrary fluctuation to which the liquid is subjected.
Consequently the TOW dynamics evolve according to a particularly simple rule: in addition to the fluctuation, if machine $k$ is played at each time $t$, $+1$ and $-\omega$ are added to $X_k(t-1)$ when rewarded and non-rewarded, respectively (Fig.~\ref{fig:stow}).
The authors have shown that this simple dynamics gains more rewards (coins or packet transmissions) than that obtained by other popular algorithms for solving the MBP~\cite{kim1,kim2}.

\subsection{The Tug-of-War Principle}

In this subsection, we derive the learning rules of the TOW dynamics from a thought experiment, so that we can obtain the nearly optimal weighting parameter $\omega_0$. 
In many popular algorithms such as $\epsilon$-greedy algorithm, an estimate for reward probability is updated only in a selected arm.
In contrast, we consider the case that the sum of the reward probabilities $\gamma$ $=$ $P_A$ $+$ $P_B$ is given in advance.
Then, we can update both estimates simultaneously as follows,

\begin{center}
\begin{tabular}{cccc}
A: & $\frac{N_A - L_A}{N_A}$ & B: & $\gamma \hspace{1mm} - \frac{N_A - L_A}{N_A}$, \\
A: & $\gamma \hspace{1mm} - \frac{N_B - L_B}{N_B}$ & B: & $\frac{N_B - L_B}{N_B}$. 
\vspace{2mm}
\end{tabular}
\end{center}

\noindent
Here, the top and bottom rows give the estimates based on $N_A$ times selecting A and $N_B$ times selecting B, respectively.

Each expected reward based on $N_A$ times selecting A and $N_B$ times selecting B is given as follows, 
\begin{eqnarray}
Q^{\prime}_k & = & N_k \hspace{1mm} \frac{N_k - L_k}{N_k} + N_j \hspace{1mm} \bigl( \gamma \hspace{1mm} - \frac{N_j - L_j}{N_j} \bigr) \nonumber \\
 & = & N_k - L_k + (\gamma - 1) \hspace{1mm} N_j + L_j . \label{eq:qAp}
\end{eqnarray}
Here, $j$ is $B$ if $k$ is $A$, or $A$ if $k$ is $B$.  
These expected rewards $Q^{\prime}_k$s are not the same as the learning rules of the TOW, $Q_k$s in Eq.(\ref{qq}).  
However, the following difference is directly used in the TOW,
\begin{equation}
Q_A - Q_B  =  (N_A - N_B) - (1 + \omega)\hspace{1mm}(L_A- L_B)\label{eq:dq}.
\end{equation}
When we transform the expected rewards $Q^{\prime}_k$s into
\begin{eqnarray}
Q^{\prime \prime}_k & = & Q^{\prime}_k / (2 - \gamma), \label{eq:qApp}
\end{eqnarray}
we can obtain the difference 
\begin{equation}
Q^{\prime \prime}_A - Q^{\prime \prime}_B  =  (N_A - N_B) - \frac{2}{2-\gamma} \hspace{1mm} (L_A - L_B). \label{eq:dqpp}
\end{equation}
Comparing the coefficient of Eq.(\ref{eq:dq}) and (\ref{eq:dqpp}), those two differences are always equal when $\omega$$=$$\omega_0$ satisfies
\begin{equation}
\omega_{0}  =  \frac{\gamma}{2-\gamma}. \label{eq:w0}
\end{equation} 
Eventually, we can obtain the nearly optimal weighting parameter $\omega_0$ in terms of $\gamma$.

This derivation means that the TOW has an equivalent learning rule with the system that is able to update both of the two estimates simultaneously.
The TOW can imitate the system that determines its next moves at time $t+1$ in referring to the estimate of each machine even if it was not selected at time $t$, as if the two machines were selected simultaneously at time $t$. 
This unique feature in the learning rule is one of origins of the high performance of the TOW.

We carried out Monte Carlo simulations and confirmed that the performance of the TOW with $\omega_0$ is comparable to its best performance, i.e., TOW with $\omega_{opt}$.
Detailed descriptions on these results will be presented elsewhere~\cite{principle}.
In addition, the essence of the process described here can be generalized to $K$-machine and $M$-player cases.
All we need is the following $\omega_0$:
\begin{eqnarray}
\omega_0  & = & \frac{\gamma^{\prime}}{2 - \gamma^{\prime}} ,\label{w0m1} \\
\gamma^{\prime} & = & P_{(M)} + P_{(M+1)} .\label{w0m2}
\end{eqnarray} 
Here, $P_{(M)}$ denotes the top $M$-th reward probability.
In fact, for $K$-machine and $M$-player cases, we have designed a physical decision-making device that achieves the overall optimal state quickly and accurately~\cite{bom}.

\section{The Tug-of-War Bombe}

\begin{figure}[b]
\centering
\includegraphics[height=45mm]{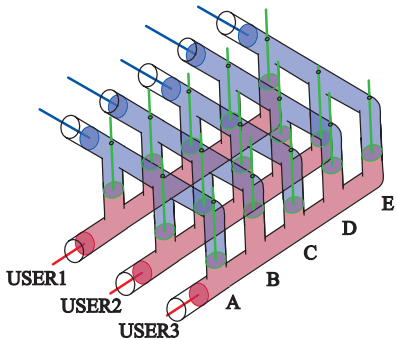}
\caption{The TOW Bombe for 3 users and 5 channels.}
\label{fig:bombe}
\end{figure}

The decision-making device called the `Tug-of-War (TOW) Bombe' for 3 users ($1, 2$, and $3$) and 5 channels ($A, B, C, D$, and $E$) is illustrated in Figure~\ref {fig:bombe}.
Two kinds of incompressible-fluid (red and blue) are filled in coupled cylinders. 
The red (bottom) fluid handles the `decision-making of a user', while the blue (upper) one handles the `interaction among users'. 
Channel selection of each user at each iteration is determined by the height of a green adjuster (a fluid interface level), and the highest channel is chosen.
When the movements of red and blue adjusters stabilize to reach equilibrium, the `tug-of-war principle' in red fluid holds for each user.
In other words, when one interface goes up, other four interfaces fall down, and efficient channel selections are attained.
Simultaneously, the `action-reaction law' is held by blue fluid (i.e., if the interface level of user1 goes up, the interface levels of user2 and 3 fall down), which contributes to avoid collisions, and the TOW Bombe is able to search for an overall optimization solution accurately and quickly.

The dynamics of the TOW Bombe are expressed as follows: 
\begin{eqnarray}
Q_{(i,k)}(t)  & = &  \Delta Q_{(i,k)}(t) + Q_{(i,k)}(t-1) \nonumber \\
 & & - \frac{1}{M-1}\sum_{j \neq i} \Delta Q_{(j,k)}(t), \\
X_{(i,k)}(t) & = & \hspace{-2mm} Q_{(i,k)}(t) - \frac{1}{N-1} \sum_{l \neq k} Q_{(i,l)}(t). 
\end{eqnarray} 
Here, $X_{(i, k)}(t) $ denotes the height of the interface of user $i$ and channel $k$ at iteration step $t$. 
If channel $k$ is chosen for user $i$ at time $t$, $\Delta Q_{(i, k)}(t)$ is $+1$ or $-\omega $ according to the result 
(rewarded or not). Otherwise, it is $0$. 

In addition to the above-mentioned dynamics, oscillations are added to $X_{(i, k)}$. 
These oscillations are given from the external by controlling the blue and red adjusters appropriately. 
In this paper, we show the cases where the completely-synchronized oscillations $osc_{(i, k)}(t) $ are added to all the users,  
\begin{equation}
osc_{(i,k)}(t) = A \hspace{1mm} sin(2 \pi t / 5 + 2 \pi (k-1)/5) .
\end{equation}
Here, $k = 1$, $\cdots$, $5$. 

Thus, the TOW Bombe operates only by adding an operation which goes up or down the interface level ($+1$ or $-\omega$) according to the result (success or failure of packet transmission) for each user (total $M$ times) at every time.
After these operations, the interface levels move according to the volume conservation law, and it calculates next selection for each user.
In the each user's selection, an efficient search is realized due to the `TOW principle' which can obtain a solution accurately and quickly in trial-and-error tasks. 
Moreover, by the interaction between users via blue fluid, the `Nash equilibrium' can be avoided consequently, and it achieves the overall optimization called `social maximum'~\cite{game}.

\section{Results}

In order to show that the TOW Bombe certainly avoids the Nash equilibrium and regularly achieves an overall optimization, 
we consider a case where ($P_A$, $P_B$, $P_C$, $P_D$, $P_E$) $=$ ($0.03$, $0.05$, $0.1$, $0.2$, $0.9$) as a typical example.
A part of the payoff tensor that has $125$ (=$5^3$) elements is described as follows for simplicity; only matrix elements for which each user does not choose low-ranking $A$ and $B$ are shown (Table~\ref {table:2}, \ref {table:3}, and \ref {table:4}). 
For each matrix element, the reward probabilities are given in the order of users 1, 2, and 3. 

\begin{table}[b]
\caption{Payoff matrix of the case where ($P_C$, $P_D$, $P_E$)$=$($0.1$, $0.2$, $0.9$), user 3 chooses $C$.}
\label{table:2}
\begin{center}
\begin{tabular}{|c|c|c|c|} \hline \hline
          & 2: C        & 2: D        & 2: E \\ \hline
1: C & $1/30$, $1/30$, $1/30$ & $.05$, $.2$, $.05$ & $.05$, $.9$, $.05$ \\ \hline
1: D & $.2$, $.05$, $.05$ & $.1$, $.1$, $.1$ & $.2$, $.9$, $.1$ {\bf SM} \\\hline
1: E & $.9$, $.05$, $.05$ & $.9$, $.2$, $.1$ {\bf SM} & $.45$, $.45$, $.1$ \\\hline
\end{tabular}
\end{center}
\end{table}
\begin{table}[h]
\caption{Payoff matrix of the case where ($P_C$, $P_D$, $P_E$)$=$($0.1$, $0.2$, $0.9$), user 3 chooses $D$.}
\label{table:3}
\begin{center}
\begin{tabular}{|c|c|c|c|} \hline \hline
          & 2: C        & 2: D        & 2: E \\ \hline
1: C & $.05$, $.05$, $.2$ & $.1$, $.1$, $.1$ & $.1$, $.9$, $.2$ {\bf SM} \\ \hline
1: D & $.1$, $.1$, $.1$ & $2/30$, $2/30$, $2/30$ & $.1$, $.9$, $.1$ \\\hline
1: E & $.9$, $.1$, $.2$ {\bf SM} & $.9$, $.1$, $.1$ & $.45$, $.45$, $.2$ \\\hline
\end{tabular}
\end{center}
\end{table}
\begin{table}[h]
\caption{Payoff matrix of the case where ($P_C$, $P_D$, $P_E$)$=$($0.1$, $0.2$, $0.9$), user 3 chooses $E$.}
\label{table:4}
\begin{center}
\begin{tabular}{|c|c|c|c|} \hline \hline
          & 2: C        & 2: D        & 2: E \\ \hline
1: C & $.05$, $.05$, $.9$ & $.1$, $.2$, $.9$ {\bf SM} & $.1$, $.45$, $.45$ \\ \hline
1: D & $.2$, $.1$, $.9$ {\bf SM} & $.1$, $.1$, $.9$ & $.2$, $.45$, $.45$ \\\hline
1: E & $.45$, $.1$, $.45$ & $.45$, $.2$, $.45$ & $.3$, $.3$, $.3$ {\bf NE} \\\hline
\end{tabular}
\end{center}
\end{table}

`Social maximum ({\bf SM})' is a state in which the maximum amount of total reward sum is obtained by all the users. 
In this problem, the social maximum corresponds to a `segregation state' in which the users choose top three different machines ($C, D, E$) respectively; there exist six segregation states that are indicated by {\bf SM} in the Tables.
On the other hand, the Nash equilibrium ({\bf NE}) is a state in which all the users choose machine $E$ independently of others' decisions; machine $E$ gives the reward with the highest probability when each user behaves in a selfish manner.

\begin{figure}[b]
\centering
\vspace{2mm}
\includegraphics[height=45mm]{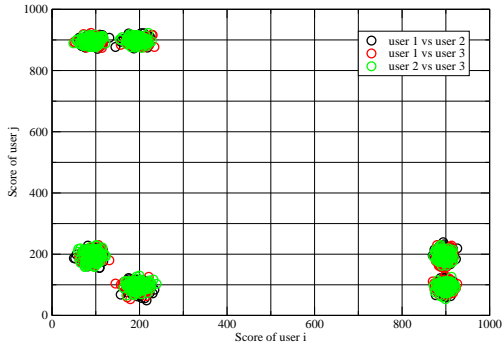}
\vspace{2mm}
\caption{Scores of the TOW Bombe in the case where ($P_A$, $P_B$, $P_C$, $P_D$, $P_E$) $=$ ($0.03$, $0.05$, $0.1$, $0.2$, $0.9$).}
\label{fig:dots}
\end{figure}

The performance of the TOW Bombe was evaluated by a score: the number of rewards (coins) a user obtained in his (her) $1000$ plays. 
In cognitive radio, the score corresponds to the amount of packets that has successfully transmitted. 
Figure~\ref {fig:dots} shows the scores of the TOW Bombe in the typical example where ($P_A$, $P_B$, $P_C$, $P_D$, $P_E$) $=$ ($0.03$, $0.05$, $0.1$, $0.2$, $0.9$). 
Since $1000$ samples were used, there are $1000$ circles for each data. 
Each circle indicates the score obtained by user $i$ (horizontal axis) and user $j$ (vertical axis) for one sample. 
There exist six clusters in Figure~\ref {fig:dots}. 
These clusters correspond to the two dimensional projections of the six segregation states, implying the overall optimization.
The social maximum points are given as follows: (the score of user $1$, the score of user $2$, the score of user $3$) $=$ ($100$, $200$, $900$), ($100$, $900$, $200$), ($200$, $100$, $900$), ($200$, $900$, $100$), ($900$, $100$, $200$), and ($900$, $200$, $100$).
The TOW Bombe did not reach the Nash equilibrium state ($300$, $300$, $300$).

\begin{figure}[t]
\centering
\includegraphics[height=45mm]{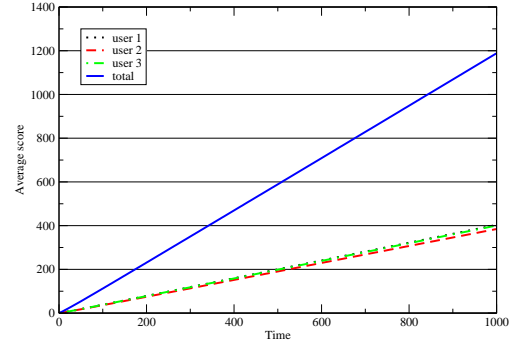}
\vspace{2mm}
\caption{Sample averages of the scores of the TOW Bombe in the case where ($P_A$, $P_B$, $P_C$, $P_D$, $P_E$) $=$ ($0.03$, $0.05$, $0.1$, $0.2$, $0.9$).}
\label{fig:trs}
\end{figure}

Figure~\ref {fig:trs} shows sample averages of the scores until $1000$ plays, where we showed the average of each user's score and that of the total score of all the users.
We can see that the average total score has gained $100$$+$$200$$+$$900$$=$$1200$, which is the value of the social maximum, while the fairness is maintained in most cases.
Here, we set parameter $\omega$ at $0.08$ (Eq. (\ref {w0m1}) and (\ref {w0m2}) were calculated as $\gamma^{\prime}$$=$$P_B$$+$$P_C$).

\section{Conclusion and Discussion}

We demonstrated that an analog decision-making device, called the TOW Bombe, is implemented physically by using two kinds of incompressible-fluid in coupled cylinders and achieves overall optimization in the channel allocation problem in cognitive radio.
The TOW Bombe enables to solve the allocation problem for $M$ users and $N$ channels by repeating $M$ up-and-down operations of the fluid interface levels in the cylinders at each iteration; it does not require the calculation of exponentially-many ($O(N^M)$) evaluation values that are required when using a conventional digital computer.
This suggests that an advantage of analog computation do exist even in today's digital age.

The TOW Bombe can also be implemented on the basis of quantum physics.
In fact, the authors have exploited optical energy transfer dynamics between quantum dots to construct the decision-making device~\cite{qdm,qdm2}. 
Our method may be applicable not only to a class problem derived from cognitive radio but also to broader varieties of game payoff matrices, implying that wider applications are expected. 
We will report these observations and results elsewhere in the future.

\section*{Acknowledgement} 

This work was partially undertaken when the authors belonged to the RIKEN Advanced Science Institute, which was reorganized and integrated into RIKEN as of the end of March, 2013. 
We thank Prof. Masahiko Hara and Dr. Etsushi Nameda for valuable discussions and advice.
We are grateful to Dr. Makoto Naruse at National Institute of Information and Communications Technology and Prof. Hirokazu Hori at University of Yamanashi for useful argument about the TOW Bombe and its quantum extension. 

\end{document}